\documentclass[letterpaper, 10 pt, conference]{ieeeconf}  
\usepackage[ansinew]{inputenc}

\IEEEoverridecommandlockouts                              

\usepackage{graphics} 
\usepackage{epsfig} 
\usepackage{amsmath} 
\usepackage{flushend}

\title{\LARGE \bf
	Improved Signed Distance Function for 2D Real-time SLAM and Accurate Localization}

\author{Xingyin Fu$^{1}$, Zheng Fang$^{2}$, Xizhen Xiao$^{1}$, Yijia He$^{1}$, Xiao Liu$^{1}$
	\thanks{$^{1}$Xingyin Fu, Xizhen Xiao, Yijia He, and Xiao Liu are with the Megvii (Face++) Technology Inc.,
		Beijing, China, {fuxingyin, xiaoxizhen, heyijia, liuxiao@megvii.com}}%
	\thanks{$^{2}$Zheng Fang is with the Faculty of Robot Science and Engineering, Northeastern University, Shenyang, China, 110169.
		{\tt\small fangzheng@mail.neu.edu.cn}}%
}

\begin{document}
	
	\maketitle
	\thispagestyle{empty}
	\pagestyle{empty}

	\begin{abstract}
		
		Accurate mapping and localization are very important for many industrial robotics applications. In this paper, we propose an improved Signed Distance Function (SDF) for both 2D SLAM and pure localization to improve the accuracy of mapping and localization. To achieve this goal, firstly we improved the back-end mapping to build a more accurate SDF map by extending the update range and building free space, etc. Secondly, to get more accurate pose estimation for the front-end, we proposed a new iterative registration method to align the current scan to the SDF submap by removing random outliers of laser scanners. Thirdly, we merged all the SDF submaps to produce an integrated SDF map for highly accurate pure localization. Experimental results show that based on the merged SDF map, a localization accuracy of a few millimeters (5mm) can be achieved globally within the map. We believe that this method is important for mobile robots working in scenarios where high localization accuracy matters.
	\end{abstract}
	
	\section{INTRODUCTION}
	Nowadays, autonomous robots are widely used in industrial and logistic areas, etc. To achieve autonomous navigation in those natural environments, the robots usually need to firstly map the environments using onboard sensors and then localize themselves in the map. For mapping the environments, Simultaneously Localization And Mapping (SLAM) is commonly used for mobile robots. Currently, many 2D mapping methods have been proposed , for instance, GMapping \cite{grisettiyz2005improving}, Cartographer \cite{hess2016real} and Hector SLAM \cite{kohlbrecher2011flexible}. However, those methods usually are grid-based methods that build maps in discrete grids. The accuracy of the map highly depends on the resolution of the grid. For memory saving issue, most time the resolution is set to 2-5 centimeters. Unfortunately, this resolution is not good enough for highly accurate mapping and it is also not good for highly accurate localization based on the map.
	
	Localization based on a known map without mapping is usually called pure localization. In the past two decays, many excellent localization methods had been proposed, for example, AMCL\cite{fox2002kld}, NDT-MCL\cite{Saarinen2013a} and Kalman Filter\cite{Teslic2010}. Among those methods, AMCL that uses KLD-sampling idea is the most widely used method for 2D pure localization. This method is very robust in various environments. However, the localization accuracy of this method is usually around 2-5cm. This is good enough for service robots but is not for industrial applications. For industrial application, mobile robots usually need millimeter accuracy to accomplish docking or manipulation.
	
	\begin{figure}[t]
		\centering
		\includegraphics[scale=0.2]{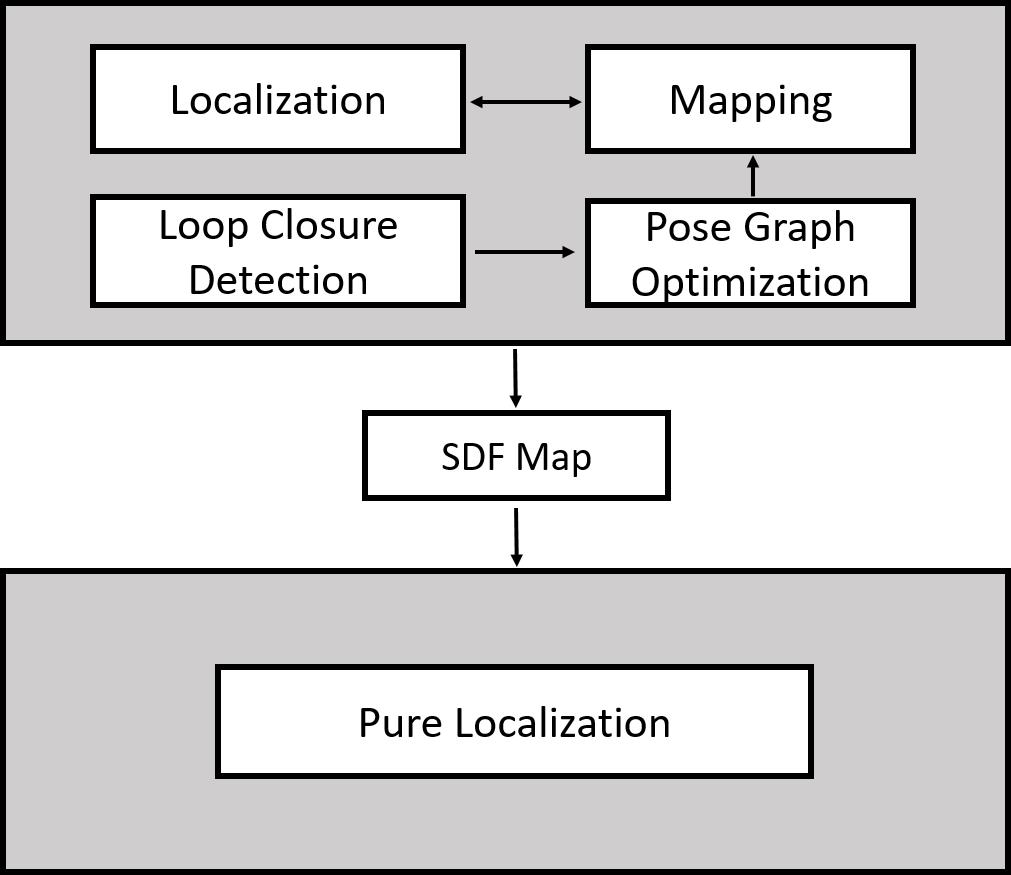}
		\caption{The same SDF map is used for localization, mapping and pure localization. }
		\label{fig:update_cells}
		\vspace{-0.2in}
	\end{figure}
	\label{sec:sdf_mapping}
	
	To achieve highly accurate mapping and localization, in this paper, we propose a new 2D laser SLAM system. In our system, an SDF map instead of an occupancy grid map is built when the robot works around in an unknown environment. 
	When the robot is working in the environment, the same SDF mapping generated is also used for pure localization.
	Experimental results show that the global localization accuracy of 5mm can be achieved, which is significantly higher compared to AMCL. Our system is also considerably more efficient. 
	
	The rest of this paper is structured as follows. Section \ref{sec:related} describes the related work. Section \ref{sec:2d_real_time_slam} presents the improved real-time 2D SLAM system. The highly accurate SDF-based pure localization is presented in Section \ref{sec:pure_localization}. Experiments are introduced in Section \ref{sec:localization} and we conclude in Section \ref{sec:conclusion}.
	\section{RELATED WORK}
	\label{sec:related}
	The SLAM system is often used to build a map of unknown environments. If a robot works in a map-based environment, a pure localization module is required. 
	Typically, the SLAM system calculates the pose by aligning the laser scan with the building map \cite{kohlbrecher2011flexible, fossel20152d, hess2016real} or the points of previous frame \cite{Censi2008}.
	Hector SLAM \cite{kohlbrecher2011flexible} and Cartographer \cite{hess2016real} build probability maps, and laser scanner pose is calculated by minimizing the probability costs of hit points. 
	To prevent the optimization get stuck in a local minimum, 
	Hector SLAM builds a multi-resolution map and uses a pyramid approach. 
	And Cartographer uses correlative scan matching \cite{olson2009real} and integrates other sensors (odometry, IMU, etc.) to pre-calculate the initial pose.
	The 2D-SDF-SLAM method \cite{fossel20152d} proposed by Fossel et al. uses Sign Distance Function (SDF) for real-time localization and mapping. 
	SDF is popularly used in 3D mapping systems, for instance, Kinectfusion \cite{newcombe2011kinectfusion, izadi2011kinectfusion}, to fuse depth received from consumer depth cameras such as Microsoft Kinect and Google Tango. 
	The 2D-SDF-SLAM method improves the mapping method to handle the situation where the laser beam is perpendicular to the surface, which is often seen in 2D laser mapping.
	We also notice that Cartographer \cite{hess2016real} implements an SDF mapping module following KinectFusion-style map updates. 
	PL-ICP \cite{Censi2008} is a variant of ICP algorithm \cite{besl1992method, rusinkiewicz2001efficient} for 2D registration. 
	The algorithm first searches for the correspondences between two sets of points.
	The pose between two frames is calculated by minimizing the point-to-line distance error of all correspondences.  
	
	Particle filters \cite{grisetti2007improved} and pose graph optimization \cite{konolige2010efficient} are two commonly used methods for the backend in 2D laser SLAM systems.
	Gmapping \cite{grisetti2007improved} and Cartographer \cite{hess2016real} are two typical systems based on particle filters and pose graph optimization. 
	Particle filter generates possible particle statuses according to the pose predicted with other sensors, such as wheel odometry. 
	And the predicted poses are evaluated by aligning laser scan with the 2D map to reduce error in system localization and mapping. 
	Cartographer reduces accumulated pose error by building submaps. 
	And loop closures are detected by matching frame points with previously built submap.
	Accumulated system tracking error is distributed with sparse pose graph optimization \cite{konolige2010efficient}.
	
	Occupancy grid map is the most commonly used map in 2D robot localization and path planning. 
	An environment is represented by fixed-sized cells,
	and the value stored in each cell represents the probability of the cell being an obstacle. 
	The occupancy map can be built with particle filter \cite{grisetti2007improved} or graph-based optimization systems \cite{kohlbrecher2011flexible, hess2016real}. 
	Currently, AMCL based pure localization using an occupancy grid is the most commonly used 2D pure localization solution. 
	However, the occupancy grid map can only represent the environment with the accuracy of map resolution,
	This may reduce the system localization and mapping accuracy.
	
	R{\"o}wek{\"a}mper et al. \cite{rowekamper2012position} achieve a few millimeters localization accuracy (5mm) at the taught-in locations by combining AMCL and PL-ICP. 
	Overall the system is localized based on AMCL.
	The PL-ICP is only used when a robot is close to a taught-in location. 
	PL-ICP calculates pose by registering the current frame with a neighbor reference keyframe. 
	Switching between AMCL and PL-ICP may cause trajectory discontinuities. 
	And the system achieves millimeters localization accuracy only when the system is close to a taught-in location. 
	Overall the system achieves 5cm localization accuracy with AMCL.
	
	\section{2D REAL-TIME SLAM}
	\label{sec:2d_real_time_slam}
	\subsection{Improved SDF-Base Mapping}
	We found that an accurate SDF map is especially important for achieving a highly accurate localization. 
	The 2D-SDF mapping method proposed by Fossel et al. \cite{fossel20152d} achieves better results in building 2D laser maps than KinectFusion-style map updates. 
	Our SDF mapping approach is built upon the 2D-SDF mapping method.
	We mainly make the following three improvements. 
	
	\subsubsection{Update SDF Map with Extended Neighbor Points}
	Since the laser scanner acquires laser points at equal angular intervals, the points near the laser center is dense, while the far points are sparse.
	The points away from the center are more conducive to improving the accuracy of the pose angle.
	We improve the 2D-SDF-SLAM method for far away point mapping. 
	\label{subsec:sdf_mapping}
	\begin{figure}[thpb]
		\centering
		\includegraphics[scale=0.15]{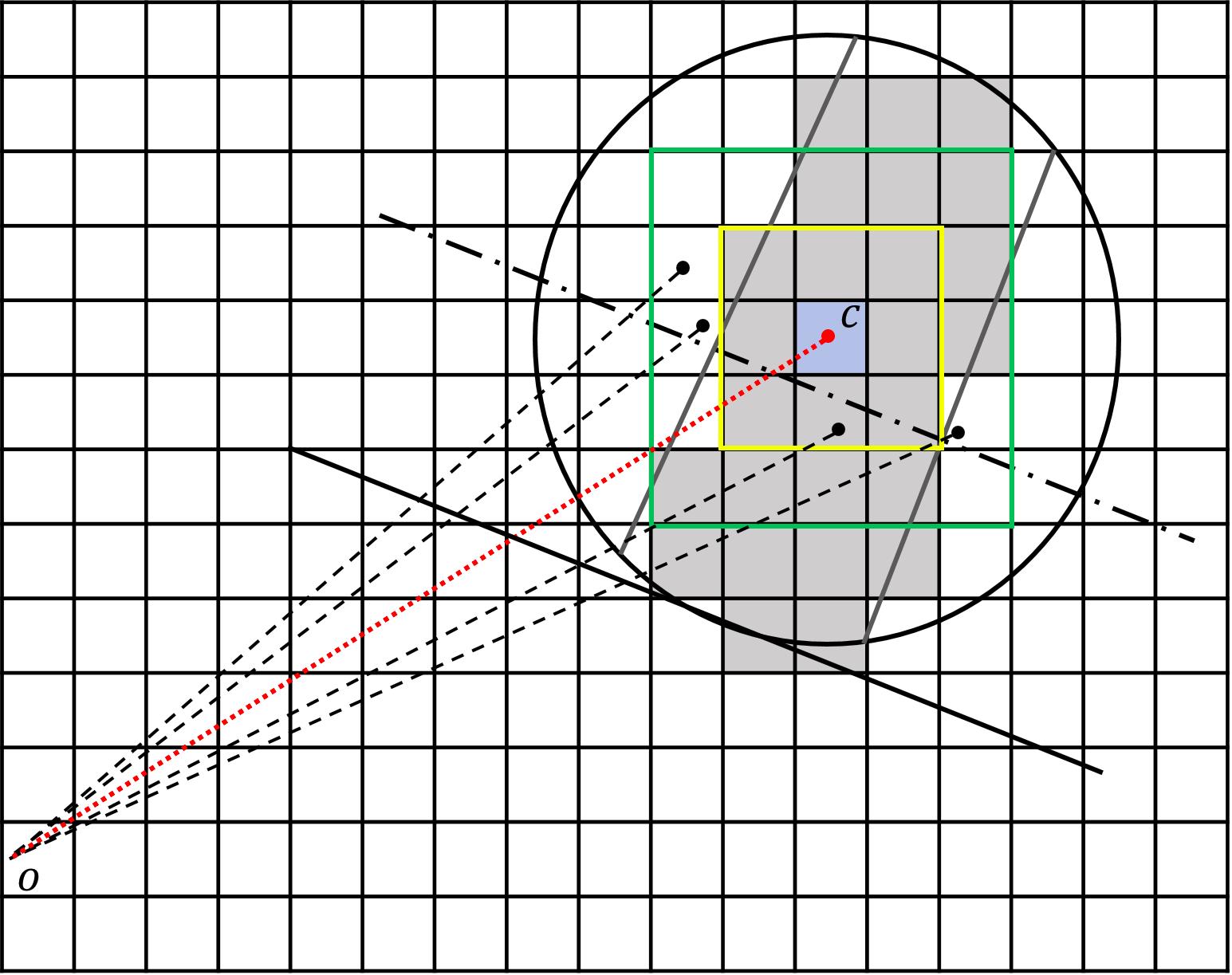}
		\caption{The search area is gradually extended to collect more points for calculating the regression line. }
		\label{fig:update_cells}
	\end{figure}
	\label{sec:sdf_mapping}
	
	To update the map with a new frame, 
	following the 2D-SDF-SLAM, we first gather the points falling in each cell and calculate the Deming-regression line \cite{deming1943statistical} using the points collected in each cell. 
	At least three points are used when calculating the regression line. 
	If the point number collected in one cell is fewer than three, 
	the points hitting in the neighbor cells are also collected.
	We gradually expand the scope to collect more points. 
	As displayed in Figure \ref{fig:update_cells}, 
	only one point fallen in the cell $c$. 
	Therefore, we expand the search area to collect more points. 
	The points hit in the nearest eight cells around the cell $c$ are also collected. 
	However, only one additional point was found in the nearest eight cells. 
	Therefore, we expanded the search area again to the green box. 
	At the second extension, a total of four points were collected. 
	The four points collected were used to compute the regression line. 
	And the regression line was used to update the SDF values of the cell $c$ and its neighbors. 
	
	To update the SDF values of neighbor cells, 
	the cells whose distance is less than $K$ are collected as indicated with the black circle in Figure \ref{fig:update_cells}. 
	$K$ is set to the truncation distance of the SDF map. 
	We project the centers of the cells enclosed in the black circle to the regression line and calculate the projection point.
	We update the cells whose projection point on the regression line falling within the range $(c_x \pm (1+0.5*\rm{e})*\rm{r}, c_y \pm (1 + 0.5*\rm{e})*\rm{r})$, 
	where $(c_x, c_y)$ indicates the 2D coordinate of the cell that causes the update,
	and $e$ is the number of times that we expands the search area. 
	$r$ is the map resolution. 
	More extensions will lead to a broader update range. 
	
	\begin{figure}[thpb]
		\centering
		\includegraphics[scale=0.25]{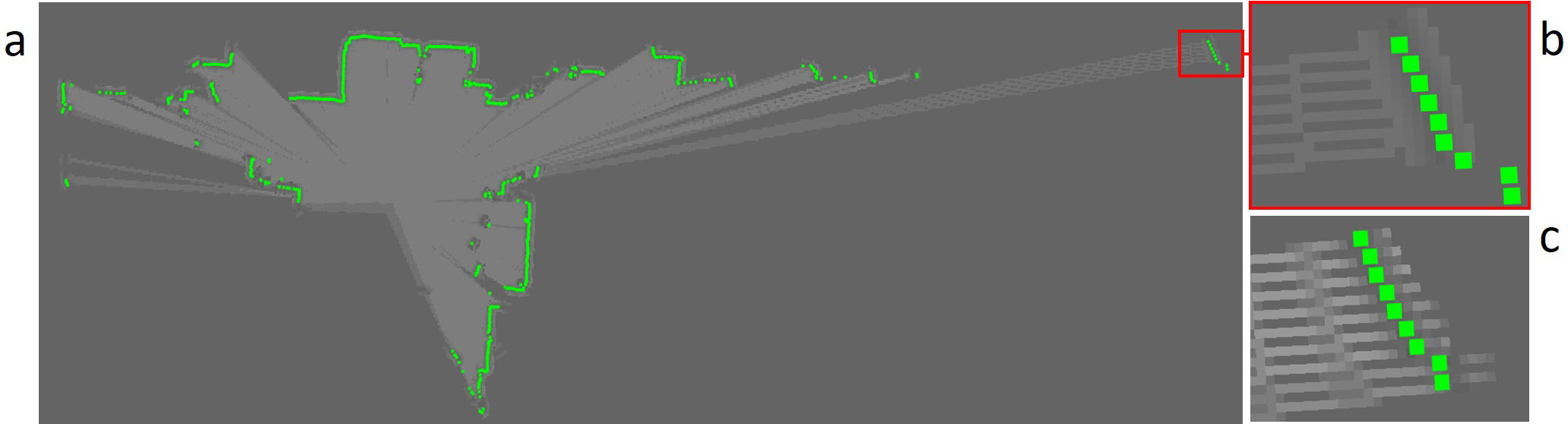}
		\caption{(a) Map update results for long distance cells. (b) is the map built with expansion and (c) is without expansion. }
		\label{fig:tsdf_mapping}
		\vspace{-0.2in}
	\end{figure}
	We found that expanding the update scope is particularly important for building high-resolution maps (for instance, 3cm or 5cm resolution). 
	And it is also crucial for updating the values of long-distance cells because of the sparse hit points as displayed in Figure \ref{fig:tsdf_mapping}. 
	As can be seen from the figure, the points displayed in the red box are far away from the laser origin, 
	The updated result of our method is displayed in (b). 
	The map is densely updated even in places where there are only sparse hit points.
	In contrast, (c) is the results updated without expansion. 
	
	In general, if less than three points are collected, the search area is only allowed to expand once for map resolution 10cm to 20cm. 
	For map resolution 5cm or less, at most three expansion times are used. 
	If less than two points are collected when the maximum expansion is reached, we will give up the updates.
	The collected points are used to calculate the regression line. 
	The SDF values of the cells are updated with the regression line as follows:
	\begin{equation}
	\begin{aligned}
	&\mathbf{F}_k(\mathbf{c}) = \frac{\mathbf{W}_{k-1}(\mathbf{c}) \mathbf{F}_{k-1}(\mathbf{c}) + \mathbf{W}_{t}(\mathbf{c}) \mathbf{F}_{t}(\mathbf{c})} {\mathbf{W}_{k-1}(\mathbf{c}) + \mathbf{W}_{t}(\mathbf{c})}, \\
	&\mathbf{W}_k(\mathbf{c}) = \rm{min}(\mathbf{W}_{k-1}(\mathbf{c})+\mathbf{W}_{t}(\mathbf{c}), \ \mathbf{W}_{max})
	\end{aligned}
	\label{equ:sdf_updating} 
	\end{equation}
	where $\mathbf{F}_{k-1}(c)$ is the previous accumulated SDF value of cell $c$, 
	and $\mathbf{W}_{k-1}(c)$ is the accumulated weight. 
	$\mathbf{F}_{k}(c)$ is the current update SDF value. 
	$\mathbf{F}_{k}(c)$ is set as the distance of the cell to the regression line. 
	If the distance from laser origin to the cell along the line normal is greater than the distance from the origin to the line, $\mathbf{F}_{k}(c)$ is set to negative, otherwise it is positive. 
	$\mathbf{W}_{k}(c)$ is the current update weight. 
	We found that there is almost no difference of using different strategies for setting the weight, 
	and simply setting the current update weight to 1 works fine. 
	We set the maximum weight $\mathbf{W}_{max}$ to 10 in our system. 
	
	\subsubsection{Update Free Space}
	
	\begin{figure}[thpb]
		\centering
		\includegraphics[scale=0.15]{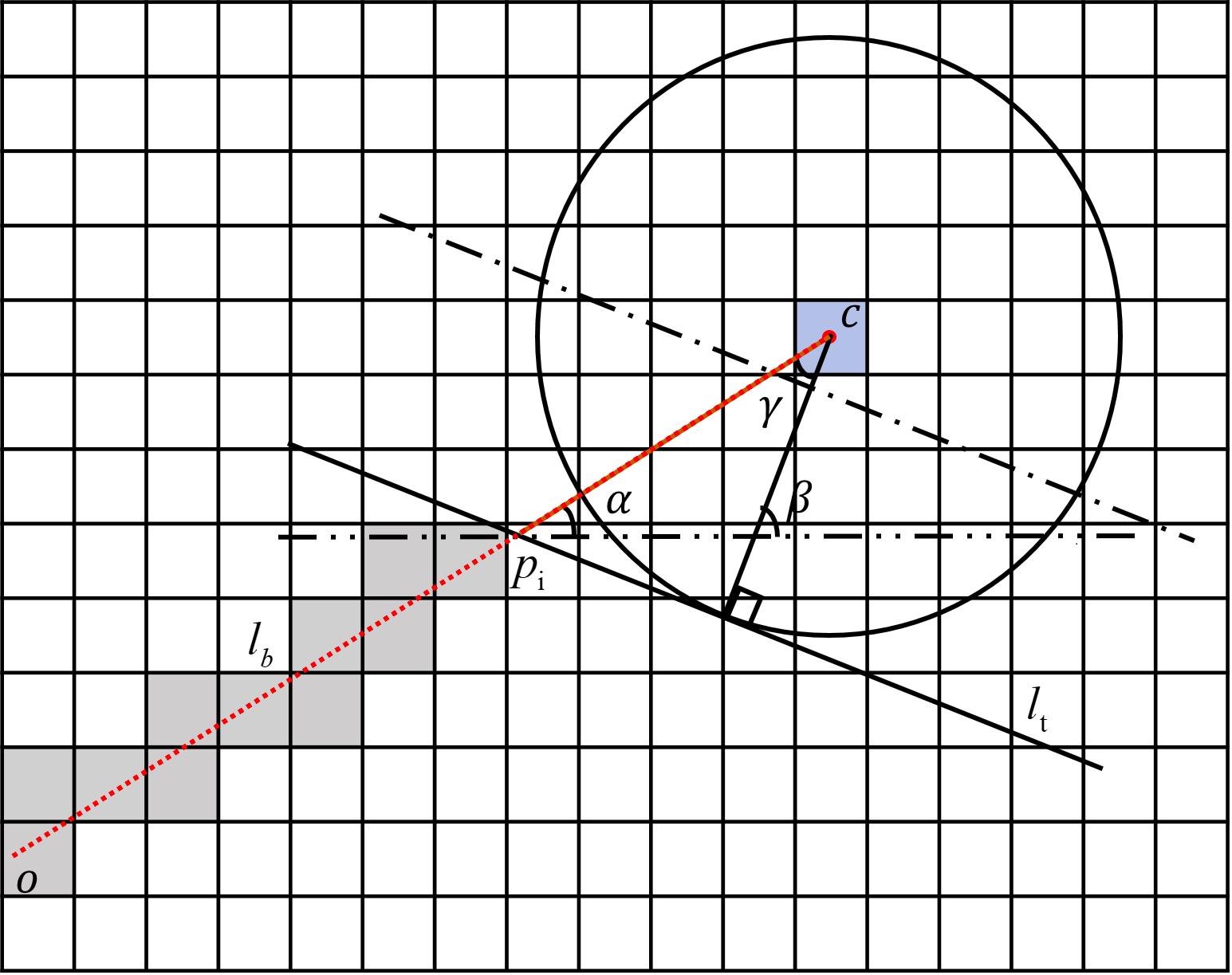}
		\caption{Update free space of an SDF map. }
		\label{update_free_points}
		\vspace{-0.2in}
	\end{figure}
	Typically, an occupancy map contains free space, unknown space and obstacle space. 
	We found that building free space is particularly important for reducing the impact of dynamic objects. 
	Dynamic objects maybe someone or a robot walking around. 
	Updated free space is resistant with dynamic scenes to arise in the SDF map, 
	which is important to reduce the outliers caused by the dynamic scene when minimizing the cost functions. 
	
	Usually, free space is updated by starting from the laser origin and marching alone a laser beam for a distance of $d_l - t_d$, where $d_l$ denotes the hit distance of a laser point and $t_d$ is the truncation distance of the SDF map. 
	Since the laser beam may not be perpendicular to the geometry surface, 
	updating cells along the line from the point $d_l$ to $d_l - t_d$ may result in incorrect free space update. 
	We propose a new method to determine the free space.
	
	As displayed in Figure \ref{update_free_points}, the gray cells are updated as free space. 
	The free space points are updated from the laser origin alone the ray to the point $p_i$, 
	where $p_i$ is the point at which laser beam $l_b$ intersects line $l_t$. 
	The point $p_i$ is calculated as follows. 
	$\alpha$ is the angle between the line $ l_b $ and the $x$ axis.
	$\beta$ is the angle between the normal of the regression line and the x-axis.
	We represent $\gamma$ as $\beta - \alpha$, and the distance between point $c$ and $p_i$ is $t_d / cos(\gamma)$.
	The free space is updated starting from the origin $o$, marching along the laser beam and ended when reach to the distance $d_b - t_d / cos(\gamma)$. 
	The update SDF of free space is directly set to the truncation distance. 
	The SDF value and weight are also fused according to Equation \ref{equ:sdf_updating}. 
	
	\subsubsection{Improve Priority Strategy}
	As presented in Section \ref{subsec:sdf_mapping}, every regression line is used to update the SDF values of neighbor cells. 
	Therefore, a cell may be updated multiple times with different regression lines.
	To determine which regression line is used to update for each cell, 
	we calculate the priority at the same time we calculate the SDF value.
	The priority is set to the distance of the cell center that needs to be updated and the cell that brought the update. 
	The cells closer to the cell that gives rise to the update are put on a higher priority. 
	
	For the update of one laser frame, a set $S_u$ is used to store the current update SDF values and priorities of all cells. 
	The cells that need to be updated are put on an update set $S_u$ one by one. 
	If the cell has been placed in the update set before, 
	we compare the current update priority and the previous one. 
	And if the current priority is higher, we replace the previously stored SDF value and priority with the current ones. 
	If the current priority and the previous one are equal, we fuse the SDF values using Equation \ref{equ:sdf_updating}, 
	and use the fused SDF value and the current priority to replace the previous values.
	If the current update has a lower priority, we give up the current update. 
	Finally, we iterate the update set and finish the one frame update with Equation \ref{equ:sdf_updating}. 
	
	\subsection{SDF-Based Lolicazation}
	\label{sec:localization}
	\subsubsection{SDF Cost Function}
	In our system, both the SLAM localization module and the pure localization module calculate pose by aligning the laser points with an SDF map. 
	The pose is calculated by minimizing the following equation: 
	\begin{equation}
	\begin{aligned}
	& \mathbf{P}^{*}=\arg \min _{\mathbf{P}} \sum_{i=0, \ldots, m} \psi \left(\rm{W}(\mathbf{c}_i) * \rm{F}(\mathbf{c}_i)\right)^{2}, \\
	& where \ \ \mathbf{c}_i = \mathbf{P} \otimes \mathbf{d}_{i}, 
	\end{aligned}
	\label{equ:calculate_pose} 
	\end{equation}
	where $\mathbf{d}_{i}$ is the 2D coordinate of laser point $i$.
	$\mathbf{P}$ is the laser pose to be estimated. 
	$\rm{F}(\mathbf{c}_i)$ is the SDF value of cell $\mathbf{c}_i$, and $\rm{W}(\mathbf{c}_i)$ is the weight. 
	We use the SDF weight in the cost function, as $\rm{W}(\mathbf{c}_i)$ indicates the confidence of the SDF value. 
	Cells that updated with more frames would be more reliable and impose a larger influence on the cost function. 
	We use Huber kernel $\psi$ to stabilize the optimization. 
	The cost function is minimized with the Gauss-Newton method. 
	
	\subsubsection{Iterative Optimization Strategy}
	\begin{figure}[thpb]
		\centering
		\includegraphics[scale=0.3]{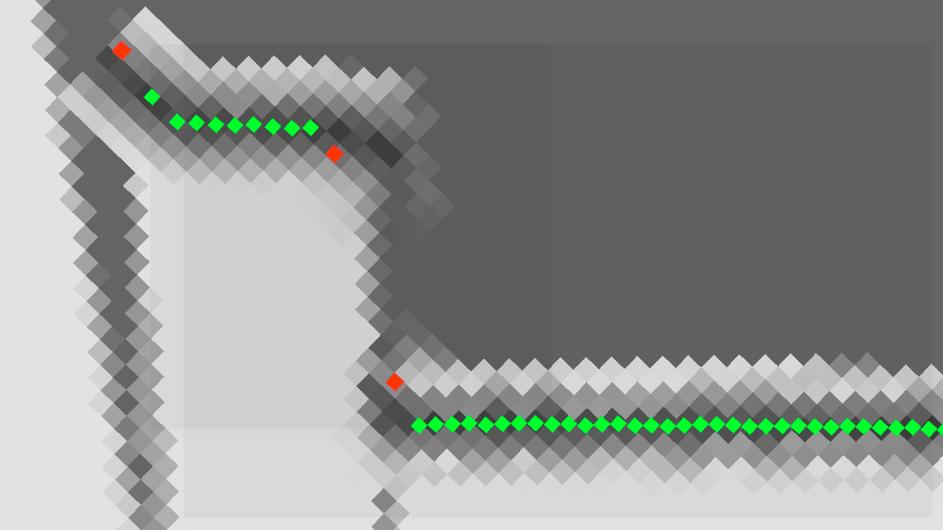}
		\caption{Outliers displayed with red points come out often in-depth discontinuity places. }
		\label{fig:outliers}
	\end{figure}
	
	Like other sensors, laser scanner takes systematic error and statistical error. 
	Take the SICK TiM5xx 2D LiDAR sensor as an example, the systematic error is within $\pm$6cm according to the technical specifications and statistical error is within $\pm$2cm.
	And we found that outliers come out often in-depth discontinuity places as displayed in Figure \ref{fig:outliers}. 
	Outliers in the free space of an SDF map do not affect the pose estimation. 
	Because the cost of hit point in the free space stays constant during the pose optimization. 
	The outliers falling inside the truncation range of the SDF map as displayed in Figure \ref{fig:outliers} would impose a large influence on the optimization. 
	Huber kernel can alleviate the impact of outliers to a certain extent. 
	We propose an iterative optimization strategy to further reduce the impact of outliers. 
	
	First, an initial pose is calculated with all the laser points by minimizing Equation \ref{equ:calculate_pose}. 
	The optimization can be initialized with the pose that calculated using a constant velocity model or predicted with other sensor data (odometry, imu, etc. ).
	Second, we minimize the cost function again with the initial pose calculated in the first step. 
	In the second step, we trim the laser points to a narrow range and discard the hit points where SDF value is greater than the threshold.
	For SICK TIM561 laser range finder, we set the truncation range to 6cm, which is identical to its systematic error. 
	Typically, in the second step, 15\% points which hit outside the truncation range are discarded in the optimization. 
	We use up to 10 iterations for the first step and up to 20 iterations for the second step. 
	And the optimization is also ceased if the error variation between two consecutive iterations falls below a threshold.
	
	\section{SDF-BASED ACCURATE PURE LOCALIZATION}
	\label{sec:pure_localization}
	We build submaps, detect loop closures and optimize the pose graph to distribute system tracking error following Cartographer. 
	In the Cartographer system, pure localization is accomplished following SLAM pipeline. 
	Laser pose is calculated by maintaining a sliding window of the newly created submaps, and the system continuously detects loop closures and optimize the pose graph. 
	We propose a new method to do pure localization. 
	When the system finishes a mapping, we merge the submaps to produce an integrated SDF map. 
	The merged map contains the regions of all submaps, and pure localization is accomplished based on the merged map. 
	We present the submap merging method as follows. 
	
	\subsection{Merge SDF Submaps}
	\label{subsec:submap_merge}
	\begin{figure}[thpb]
		\centering
		\includegraphics[scale=0.25]{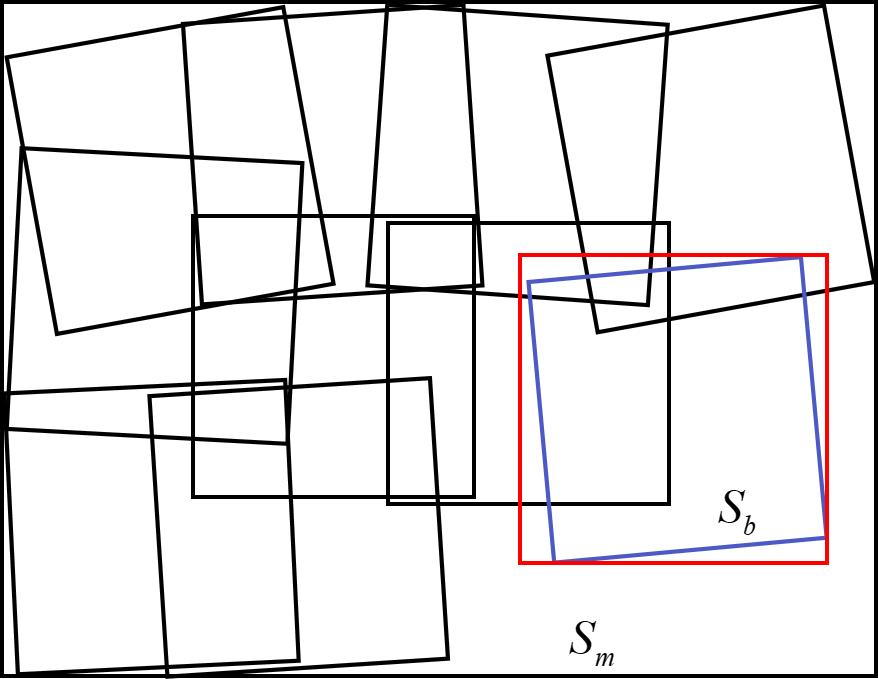}
		\caption{All submaps are merged to produce an integral map. }
		\label{fig:merge_submap}
	\end{figure}
	
	As displayed in Figure \ref{fig:merge_submap}, the inside black boxes are generated SDF submaps. 
	Due to the backend pose graph optimization, the submap boxes are not horizontal or vertical to each other as they were when initialized.
	To merge all submaps, we first calculate the position and size of the large bounding box $b_s$ as shown by the outermost one. 
	The large bounding box can cover the area of all submaps. 
	For each submap, the four corner coordinates are calculated using the optimized pose. 
	And the position and size of the enclosed bounding box $b_s$ is initialized according to the maximum and minimum corner coordinates of the submaps. 
	The merged SDF map $S_m$ is created based on the center coordinates and area of the bounding box $b_s$. 
	And all submaps are integrated into the map $S_m$. 
	
	For each submap, for instance, the blue one $S_b$ shown in Figure \ref{fig:merge_submap}, 
	we first calculate its corresponding region in the large map $S_m$ as indicated with the red box. 
	Then, we iterate the cells of the map $S_m$ in the red box and fuse the SDF values of the map $S_m$ and submap $S_b$. 
	For each cell $c_m$ in the red box of the large map $S_m$, 
	we calculate its corresponding point $p_c$ in the submap $S_b$. 
	The SDF value at $p_c$ of $S_b$ is bicubic interpolated using the values of neighbor cells. 
	The SDF value of the cell $c_m$ in the integrated map $S_m$ and the value at the corresponding coordinate $p_c$ in the blue submap are fused according to Equation \ref{equ:sdf_fusion}, 
	\begin{equation}
	\begin{aligned}
	&\mathbf{F}(\mathbf{c}) = \frac{\mathbf{W}_{m}(\mathbf{c}_m) \mathbf{F}_{m}(\mathbf{c}_m) + \mathbf{W}_{b}(\mathbf{p}_c) \mathbf{F}_{b}(\mathbf{p}_c)} {\mathbf{W}_{m}(\mathbf{c}_m) + \mathbf{W}_{b}(\mathbf{p}_c)}, \\
	&\mathbf{W}(\mathbf{c}) = \rm{max}(\mathbf{W}_{b}(\mathbf{p}_c), \ \mathbf{W}_{m}(\mathbf{c}_m)), 
	\end{aligned}
	\label{equ:sdf_fusion} 
	\end{equation}
	where $\mathbf{W}_m(\mathbf{c}_m)$ and $\mathbf{F}_m(\mathbf{c}_m)$ is previously accumulated weight and value of cell $\mathbf{c}$,  
	$\mathbf{W}_{b}(\mathbf{p}_c)$ and $\mathbf{F}_{b}(\mathbf{p}_c)$ is the weight and value of point $\mathbf{p}_c$ in submap $S_b$. 
	As submaps may overlap, the cell in map $S_m$ may have multiple correspondences in submaps. 
	The submaps are gradually integrated into the large map $S_m$, and SDF values are continually fused. 
	And for the fused weight, we set the weight to the maximum value of the corresponding points in all submaps. 
	
	\subsection{Pure Localization Based on Merged Submap}
	\label{subsec:submap_merge}
	The merged submap is used to do pure localization. 
	The initial pose can be relocalized with other sensors, such as a visual camera. 
	In the pure localization module, we register laser points with the merged map without building new submap.
	Laser pose is calculated following the pipeline of localization as introduced in Section \ref{sec:localization}. 
	Since there is no map building when pure localization is done, and usually only a few iterations (typically 5) are utilized when minimizing the cost function, we consume much less computing resources compared to the pure localization method of Cartographer. 
	Pure localization accuracy and computing consumption are discussed in Section \ref{sec:experiments}.
	
	\begin{table*}[htb]
		\caption{Trajectory evaluation results based on MIT Stata Center dataset \cite{fallon2013stata}} 
		\vspace{-0.1in}
		\label{tab:evaluate_trajectories}
		\begin{center}
			\begin{tabular}{|c|c|c|c|c|}
				\hline
				& Hector SLAM \cite{kohlbrecher2011flexible} & Cartographer Probability Grid \cite{hess2016real} & Cartographer SDF \cite{hess2016real} & Our System \\
				\hline
				2012-01-27-07-37-01\_part\_1 & 0.0114 & 0.0098 & 0.0102 & \textbf{0.0096} \\
				\hline
				2012-01-27-07-37-01\_part\_3 & 0.0121 & 0.0101 & 0.0101 & \textbf{0.0097} \\
				\hline
				2012-01-28-11-12-01 & 0.0616 & 0.0109 & 0.0106 & \textbf{0.0102} \\
				\hline
				2012-04-02-10-54-41 & 0.4538 & 0.0265 & 0.0279 & \textbf{0.0262} \\
				\hline
				2012-02-02-10-44-08 & - & 0.0141 & 0.1466 & \textbf{0.0140} \\
				\hline
				2012-05-01-12-12-25\_part\_2 & 0.0122 & 0.0100 & 0.0100 & \textbf{0.0098} \\
				\hline
				2012-05-02-06-23-02 & 0.0442 & 0.0221 & 0.0126 & \textbf{0.0123} \\ 
				\hline
				2012-01-28-12-38-24\_part\_1 & 0.0956 & 0.0118 & 0.0105 & \textbf{0.0101} \\
				\hline
				2012-01-28-12-38-24\_part\_4 & - & 0.0113 & 0.0112 & \textbf{0.0107} \\
				\hline
				2012-01-28-11-12-01 & 0.0616 & 0.0109 & 0.0105 & \textbf{0.0102} \\
				\hline
			\end{tabular} \vspace{-0.1in}
		\end{center}
	\end{table*}
	
	\section{EXPERIMENTS}
	\label{sec:experiments}
	We evaluated the trajectories generated by our systems and other state-of-the-art 2D-SLAM systems, including Hector SLAM \cite{kohlbrecher2011flexible} and Cartographer \cite{hess2016real} based on the MIT Stata Center dataset \cite{fallon2013stata}.
	Since Cartographer implements two mapping method of probability grid and SDF,
	We evaluate the performance of both.
	We also evaluate the position error to determine the accuracy that a robot localizes itself in a known map. 
	The position error is the joint error of the pure localization and robot position systems as a whole. 
	\subsection{Evaluate Odometry Accuracy}
	\label{subsec:submap_merge}
	We turn off the loop closure detection and pose graph optimization modules when generating the trajectories, and use Root Mean Square Error (RMSE) to evaluate the accuracy of the trajectories. 
	Because Hector SLAM is an odometry system and RMSE can better reflect the performance of the localization and mapping modules that we focus on. 
	The evaluation results are presented in Table \ref{tab:evaluate_trajectories}. 
	As can be seen from the table, our system performs better than the other three systems. 
	Our SDF mapping method represents the surface more accurately compared to the occupancy grid. 
	Because the surface that crosses the SDF map from positive to negative can be accurately located.
	In the occupancy grid map, the surface can only be located in the accuracy of map resolution.
	More importantly, the localization based on the SDF map has a wider convergence range than the occupancy map. 
	Optimization can find the direction of convergence within the truncation distance.
	This is a key factor that SDF map can be used robustly for pure localization. 
	
\begin{figure}[thpb]
		\centering
		\includegraphics[scale=0.35]{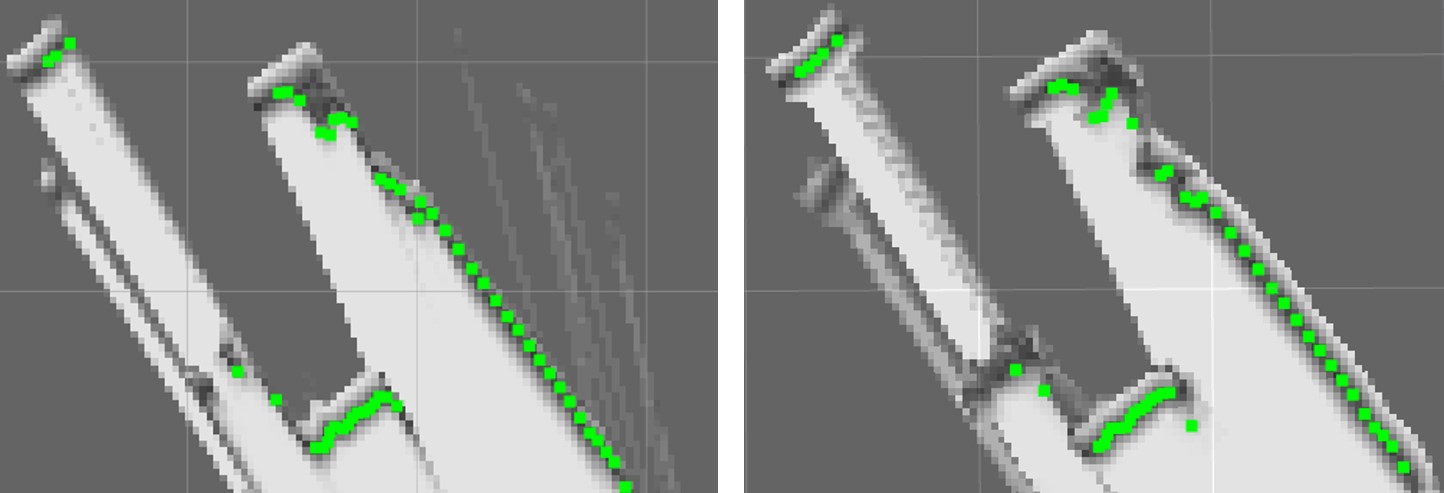}
		\caption{A comparison of the SDF maps generated by Cartographer and our system. The one on the left is generated by Cartographer \cite{hess2016real} and the one on the right is by our system.}
		\label{fig:tsdf_mapping_compare}
		\vspace{-0.2in}
	\end{figure}

	Cartographer implements SDF mapping following a KinectFusion-style map update. 
	We also compare the performance of Cartographer SDF mapping method with ours. 
	During the evaluation, we use the same parameters, for example, the same number of points used in pose solving and the same truncation distance used in the mapping.
	As displayed in Table \ref{tab:evaluate_trajectories}, our system achieves better results than Cartographer. 
	When the laser beam is closely parallel to the geometric surface, the SDF map generated by the Cartographer is not accurate, as shown in Figure \ref{fig:tsdf_mapping_compare}.
	Only a very narrow area alone the geometry surface was updated by Cartographer. 
	In contrast, we build an more accurate SDF map. 
	On the other hand, our iterative optimization strategy can trim outliers to some extent.
	
	\subsection{Evaluation the Pure Localization Accuracy}
	\label{subsec:submap_merge}
\begin{figure}[thpb]
		\centering
		\includegraphics[scale=0.2]{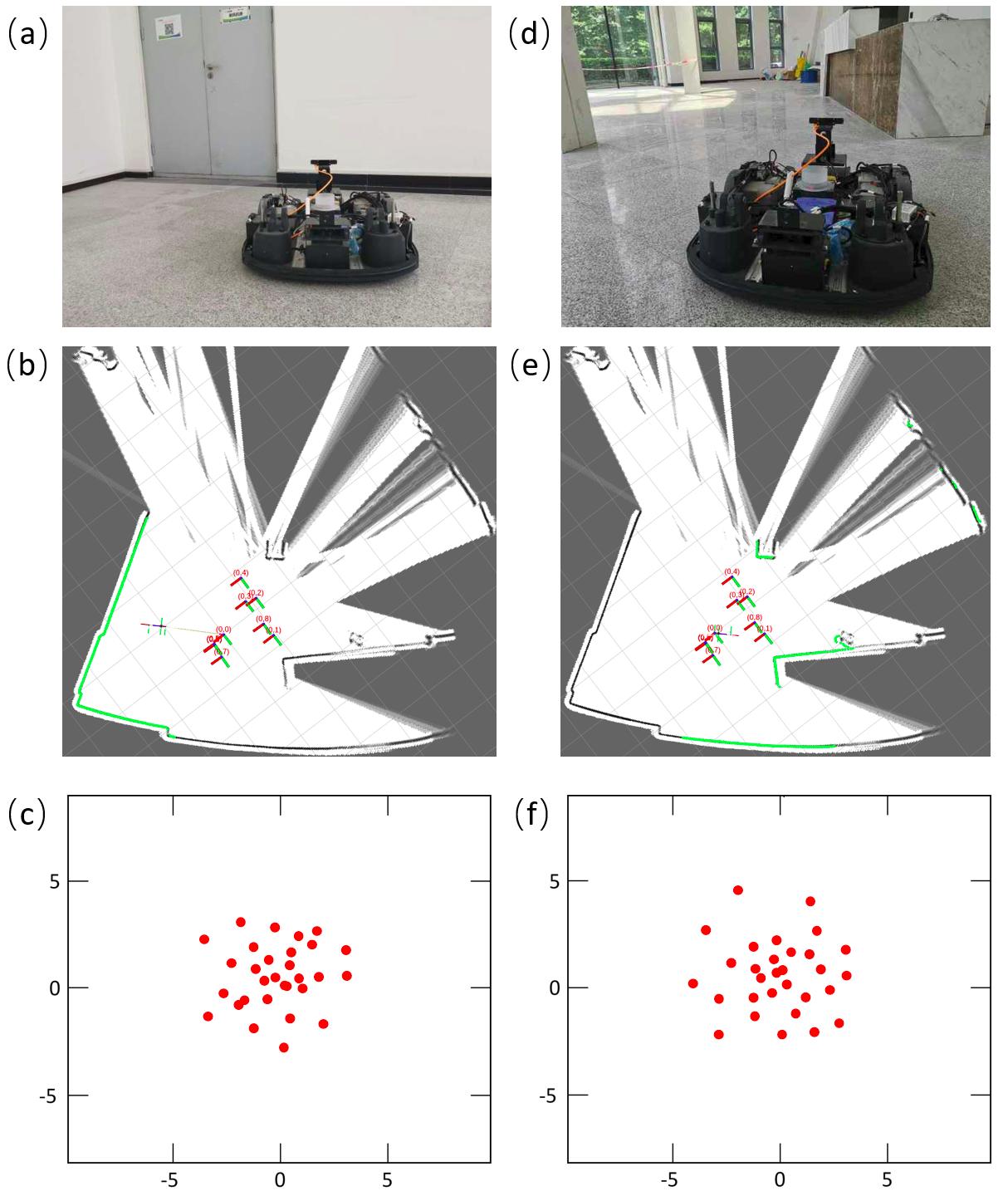}
		\caption{A robot moves from two designated positions in the map from one to the other as displayed in (a) and (d). (b) and (e) is the SDF map when the robot stands still at the two position. (c) and (f) is the robot position error at the two positions. }
		\label{fig:robto_position_expriments}
		\vspace{-0.2in}
	\end{figure}

	We evaluate the robot position error, which is measured relative to a reference position following \cite{rowekamper2012position}. 
	We arbitrarily specify two positions in the SDF map and let the robot controller control the robot to move from one position to another cyclically for reciprocation motion.
	The robot position is calculated relative to a fixed coordinate system using an April tag every time the robot was standing at a designated position. 
	We summarize 30 locations and use the deviation of each location from the center as the position error.
	The evaluation results are presented in Figure \ref{fig:robto_position_expriments}.
	As can be seen from the figure, 
	our system achieves a position error of $\pm$5mm. 
	There are always people moving during the robot position accuracy evaluation experiments. 
	And we found that small dynamic objects, such as few people or robots moving around, have little affects on robot position accuracy. In contrast, the system proposed by Röwekämper et al. \cite{rowekamper2012position} achieves $\pm$5mm accuracy only in the taught-in locations using PL-ICP method \cite{Censi2008}. 
	Overall the system can only achieve 2 $\sim$ 5cm position accuracy with AMCL. 
	
	There are two reasons why our system performs better than AMCL. 
	On the one hand, our system uses SDF map while AMCL using occupancy map. SDF map is more accurate as geometry surface can be precisely localized.
	One the other hand, we use optimization to find the optimal solution which is more accurate than particle filter used by AMCL. 
	
\begin{figure}[thpb]
		\centering
		\includegraphics[scale=0.15]{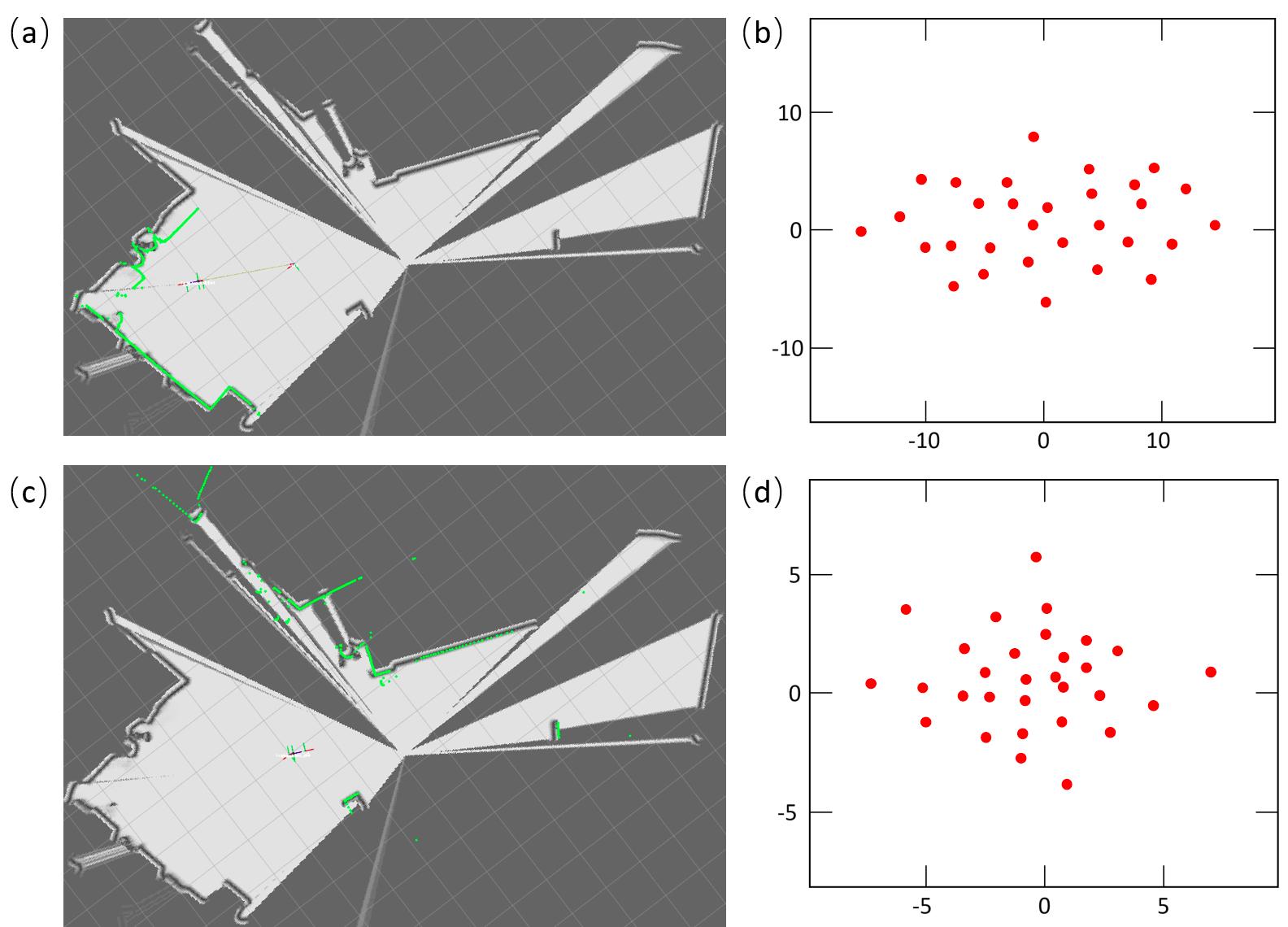}
		\caption{Evaluate the robot position accuracy in highly dynamic environment.}
		\label{fig:robto_position_with_large_dynamics}
		\vspace{-0.2in}
	\end{figure}

	We also evaluate the accuracy of robot position in a highly dynamic environment, as displayed in Figure \ref{fig:robto_position_with_large_dynamics}. 
	The environment was changed greatly, resulting in the large misalignment of the laser points and the SDF map. 
	As can be seen from the figure, the robot position error is $\pm$2cm.
	During all the experiments, our system works properly. 
	And we found that our system achieves high robustness owning to the large convergence range of SDF map. 
	
	\subsection{Compare the Pure Localization Efficiency}
	\label{subsec:submap_merge}
	We also compare the pure localization efficiency of our system with AMCL. 
	Both the AMCL and our system are executed in an NVIDIA Jetson TX2 platform and evaluated in the same environment. 
	AMCL uses the occupancy map generated by our system. 
	And both the pure localization modules run in CPU with a single thread.
	The statistical execution time of one frame is displayed in Table \ref{tab:execution_time}.
	As can be seen from the table, our system is about three times faster than AMCL.
	Pure localization efficiency is critical for multitasking systems running on embedded CPUs.
	\begin{table}[htpb]
		\caption{Compare the efficiency of AMCL and our system. }\vspace{-0.1in}
		\label{tab:execution_time}
		\begin{center}
			\begin{tabular}{|c|c|c|c|c|}
				\hline
				& median & mean & max & std \\
				\hline
				AMCL & 0.0314 & 0.0351 & 0.1050 & 0.0262 \\
				\hline			
				Our system & \textbf{0.0078} & \textbf{0.0090} & \textbf{0.0345} &  \textbf{0.0048} \\
				\hline
			\end{tabular}
		\end{center} \vspace{-0.2in}
	\end{table}
	
	\section{CONCLUSION}
	\label{sec:conclusion}
	In this paper, we improve the 2D-SDF-SLAM system and propose a new pure localization method.
	We improve the 2D-SDF-SLAM method to build a more accurate SDF map.
	To resolve random outliers of laser scanners, we propose a new iterative registration method. 
	Using fewer outliers can greatly improve localization accuracy. 
	The generated SDF map is also used for pure localization. 
	Globally, a few millimeters (5 mm) of positioning accuracy can be achieved within the map while using much less computation resources.
	This accuracy allows the creation of highly demanding robotic applications, for example, precise maneuvering tasks for docking maneuvers or movements.
	
	\addtolength{\textheight}{-12cm}   
	


	
	
	
	
	\bibliographystyle{IEEEtran}
	\bibliography{egbib}

\end{document}